\documentclass[a4paper,10pt,twocolumn]{article}  

\usepackage{geometry}
\usepackage[pdftex]{graphicx} 
\usepackage{titlesec}
\usepackage[labelfont=bf]{caption}
\usepackage{enumitem}
\usepackage{hyperref}
\usepackage{helvet}
\usepackage{etoolbox}

\DeclareCaptionFormat{myformat}{\fontsize{9}{9}\selectfont#1#2#3}
\titleformat{\section}
  {\normalfont\fontsize{10}{0}\bfseries}{\thesection}{0em}{}
\titlespacing\section{0pt}{12pt }{3pt }
\newcommand\Authfont{\fontsize{14}{0}\selectfont\sffamily}
\newcommand\Affilfont{\fontsize{12}{0}\itshape\selectfont\sffamily}

\makeatletter
\patchcmd{\@maketitle}{  \@title}{\fontsize{14}{0}\selectfont\sffamily \bfseries \@title}{}{}

\def\bstctlcite#1{\@bsphack
  \@for\@citeb:=#1\do{%
    \edef\@citeb{\expandafter\@firstofone\@citeb}%
    \if@filesw\immediate\write\@auxout{\string\citation{\@citeb}}\fi}%
  \@esphack}
  
\makeatother

\title{Affordable Mobile-based Simulator for Robotic Surgery \vspace{-0.4cm}}

\author{
\begin{tabular}[t]{c@{\extracolsep{2em}}c@{\extracolsep{2em}}c}
\multicolumn{3}{c}{\Authfont{Piyamate Wisanuvej\textsuperscript{1,2,3}, Petros Giataganas\textsuperscript{1}, Paul Riordan\textsuperscript{1},}} \\
\multicolumn{3}{c}{\Authfont{Jean Nehme\textsuperscript{1}, and Danail Stoyanov\textsuperscript{1,2}}} \\[3mm]
\Affilfont{\textsuperscript{1}Digital Surgery Limited} & \Affilfont{\textsuperscript{2}University College London} & \Affilfont{\textsuperscript{3}Kasetsart University} \\
\Affilfont{London, UK} & \Affilfont{London, UK} & \Affilfont{Bangkok, Thailand} \\
\Affilfont{piyamate@touchsurgery.com} & \Affilfont{danail.stoyanov@ucl.ac.uk} & \Affilfont{fengpmw@ku.ac.th} \\ \vspace{-0.5cm}
\end{tabular}
}

\begin{document}
\bstctlcite{IeeeBstCtl}

\date{}
\maketitle
\thispagestyle{empty}
\pagestyle{empty}

\section*{INTRODUCTION}

Robotic surgery presents great potentials towards safer, more accurate and consistent minimally invasive surgery (MIS) \cite{Taylor2016}. However, their adoption is fundamentally dependent on the access to training facilities and extensive surgical training \cite{AlimViceFior20118a,Sridhar2017}. Robotic instruments require different dexterity skills compared to open or laparoscopic surgery as well as across different robotic systems. Surgeons, therefore, are required to invest significant time by attending extensive robotic training programs. Hands-on experiences, also, represent an additional operational cost for hospitals as the availability of robotic systems for training purposes is limited. All these technological and financial barriers for surgeons and hospitals hinder the adoption of robotic surgery technology. 

Currently, the robotic surgery scene is represented by the state-of-the-art robotic MIS (RMIS) system, the da Vinci\textregistered$\,$ surgical system (Intuitive Surgical, CA). Due to the increased interest for the system, various surgical simulators for the system have been developed over the years. They offer a computer-generated reproduction of real-world surgical procedures and surgical tasks for different levels of expertise. 
These platforms are mainly stand-alone and do not compromise patient's safety for training. However, they cost tens of thousands of dollars as they are based on either expensive, but accurate, electromagnetic systems or room-based visual tracking systems or high-end robotic manipulators. They are non-portable and often require dedicated training spaces, and are developed exclusively for the da Vinci systems. Similar limitations existed in laparoscopic surgical training but with technological advances, nowadays, low cost alternatives exist for basic surgical tasks with full performance analytics and support of generic laparoscopic instruments with cost of less than $\$$1000. These simulators are often called `take-home' simulators that surgeons can use to train anywhere and are potentially significant for countries where surgical training tools are limited.    

In this work, we present a low-cost, fully wireless, and portable solution to train basic dexterity skills for introductory-level robotic surgery. The platform can facilitate the training of basic gestures and improve the users familiarity with the hand-motor axis control and ergonomics needed for manipulating robotic instruments. It is intended for surgeons without robotic surgery training to become familiarised with new dexterity skills required. To our knowledge, this is the first attempt to demonstrate such a system for RMIS.

\begin{figure}
    \centering
    \includegraphics[width=0.9\columnwidth]{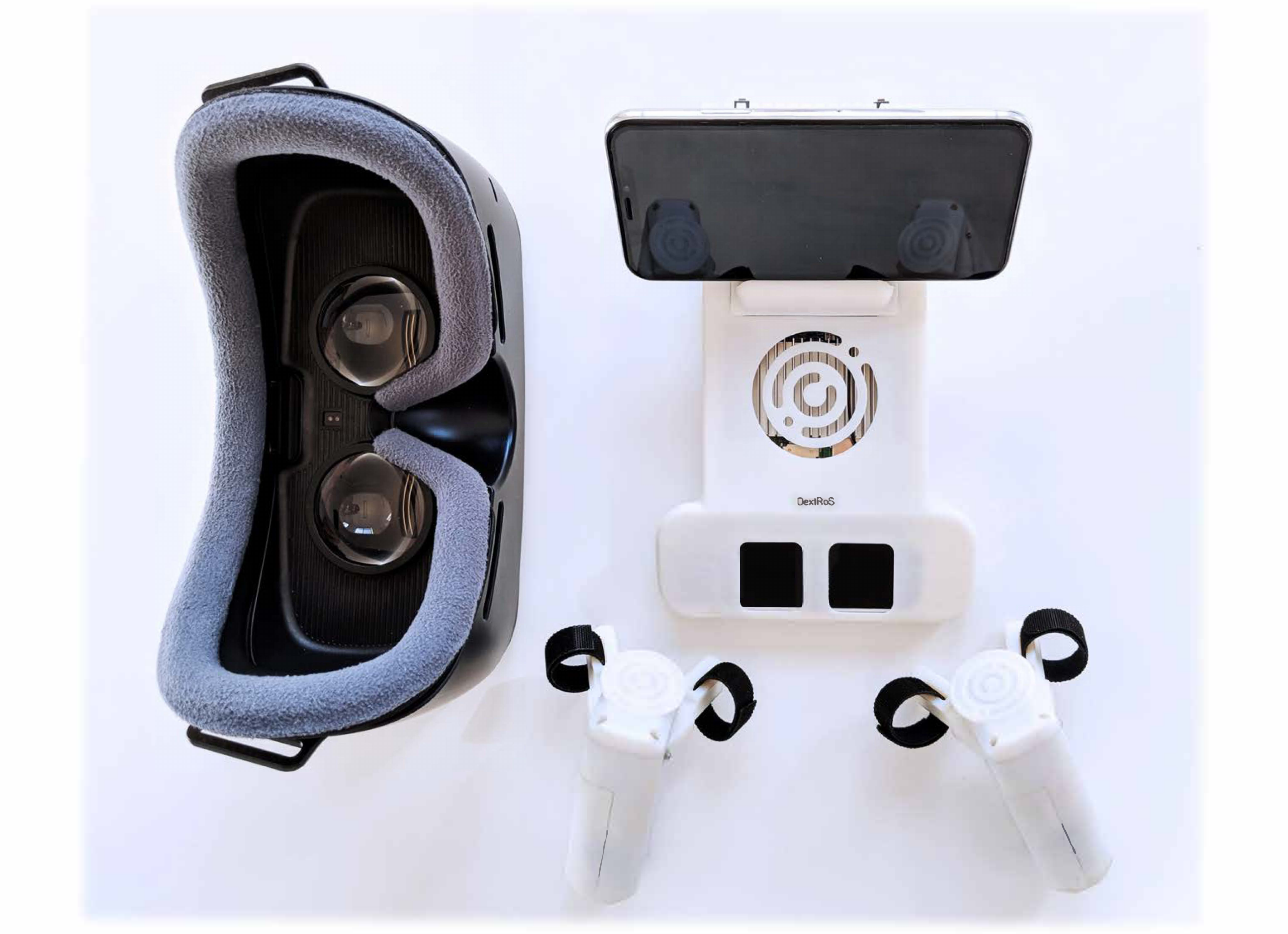}
    \caption{Prototype demonstration of the mobile surgical simulator presenting all the elements of the system. \vspace{-0.5cm}}
    \label{fig:schematic}
\end{figure}

\section*{MATERIALS AND METHODS}

\begin{figure}
      \centering
     \includegraphics[width=\columnwidth]{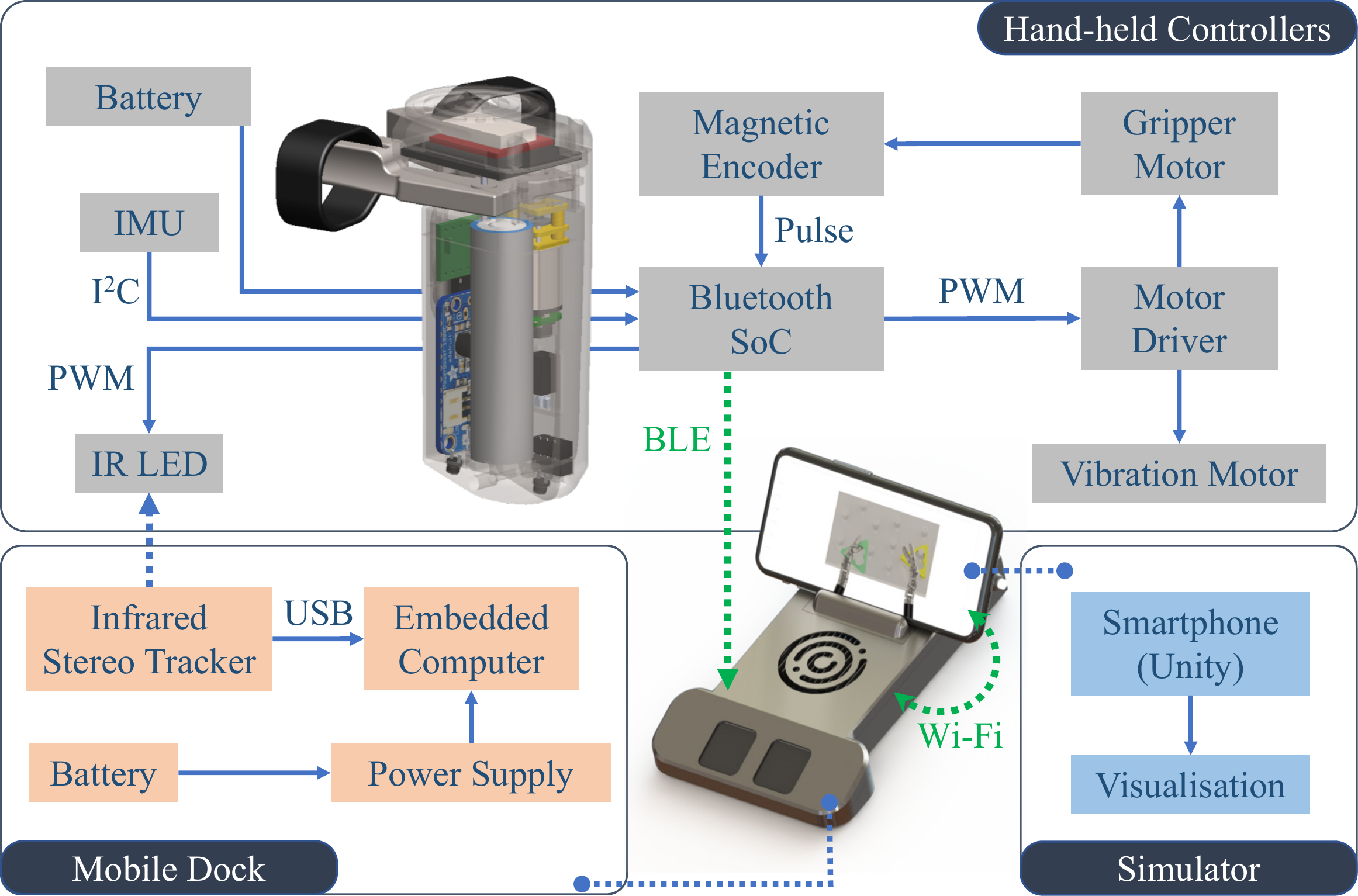}
      \caption{Overview of the system architecture along the various communications between the components.}
      \label{fig:systemOverview}
      \vspace{-0.4cm}
   \end{figure}

The proposed portable system consists of three core components as shown in Fig.~\ref{fig:schematic}: a pair of wireless hand-held master controllers with haptic and tactile feedback, a mobile dock, and a smartphone or tablet that run the simulation software. The smartphone can also housed inside a virtual reality (VR) headset for 3D visualisation. An overview of all the system components along with the communication links is presented in Fig.~\ref{fig:systemOverview}. Unlike many VR headsets, the system is self-contained and requires no external host computer. The material cost to build this system is approximately \textdollar500, which is significantly lower than existing commercial solutions.

The platform combines an inertial measurement unit (IMU) and an infrared (IR) tracking system to provide a low cost alternative to expensive commercial tracking solutions, \textit{e.g.} electromagnetic tracking. The IMU is used only for the orientation tracking as the integration errors hinder its use for position estimation. The position of each wireless hand-held controller is tracked externally using a low cost IR stereo tracker (Leap Motion, USA) that is situated in the front of a docking station and an IR LED attached at the bottom of each controller. Active trackers were preferred due to their higher contrast and robustness in detection; passive markers are more prone to false positives due to flood illumination. A simple binary thresholding scheme is adopted to detect each marker. Then, the pair of corresponding markers in the stereo images are triangulated using the calibrated cameras and the triangulated points are smoothed over time using a moving average filter. Each controller is equipped with only one LED to reduce the computational load on the embedded computer. The position data from the cameras is complemented by the orientation data from the IMU to obtain full 6-DoF tracking.

Every hand-held controller has a multifunctional button. When pressed, then the user can activate the clutch of each surgical instrument. If the buttons on both controllers are pressed, then the user can adjust the camera view by moving both controllers in relative to each other for rotation and translation.

\begin{figure}
      \centering
     \includegraphics[width=\columnwidth]{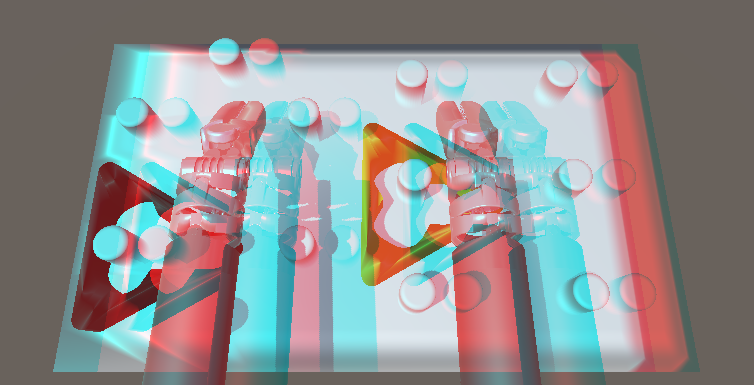}
      \caption{Stereoscopic scene of the peg transfer task for the usability experiment in anaglyph 3D format.}
      \label{fig:user_experiment_3d_view}
\end{figure}

The simulation environment used with the smartphone or tablet is built using the Unity engine (Unity Technologies, USA), while the physical interactions are simulated using the built-in physics engine (Nvidia PhysX). The articulated instruments' joint angles are calculated using a commercially available inverse kinematics (IK) solver (Rootmotion). In the presented work, a simple but well-established pick and place surgical task is demonstrated,depicted in Fig.~\ref{fig:user_experiment_3d_view}, which is an adaptation of the Fundamentals of Laparoscopic Surgery (FLS) model training task.

\section*{RESULTS}
A preliminary study was conducted with five novice users to evaluate the usability of the mobile simulator. They were asked to perform a standard peg transfer task using the platform. Each user had five minutes to get familiarised with the task, then, the actual experiment commences. The task is to perform bi-manual peg transfers from one side to the other and repeated the steps until three minutes have passed. The experiment repeated three times for each user with a minute of break in between.

\begin{figure}
\vspace{-0.4cm}
      \centering
     \includegraphics[width=\columnwidth,trim={1.4cm 0 1.3cm 0},clip]{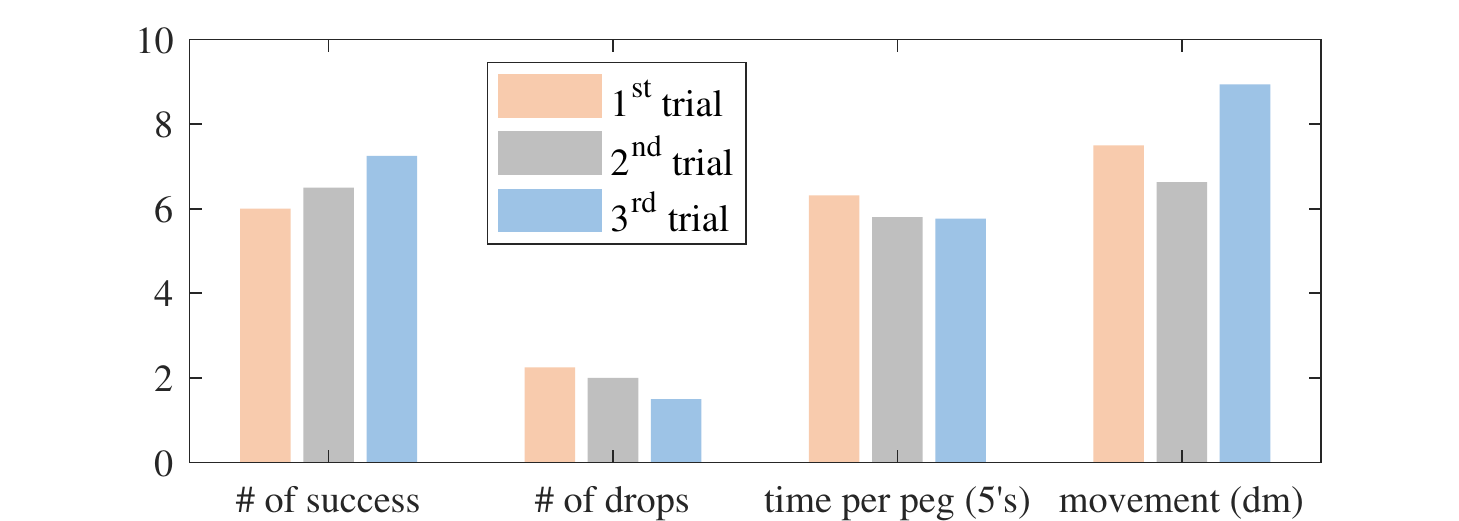}
      \caption{Results from the user experiment, showing mean values of four performance metrics among all users.}
      \label{fig:user_experiment}
\vspace{-0.4cm}
\end{figure}

The metrics measured during the experiment include: number of success transfers, number of accidental drops, time spent per transfer, and the total distance moved. Fig.~\ref{fig:user_experiment} summarises the results.

\section*{DISCUSSION}
The preliminary user study presents improvement in performance over time as indicated by the increasing number of successful transfers and the decreasing number of drops. The performance improvement of the second and the third trial over the first one is 8\% and 21\% for the number of successful transfers, and 11\% and 33\% for the number of accidental drops respectively. This suggests that the users could learn and improve their dexterity skills with the robotic simulator over a short period of time.

One common user feedback is the lack of depth perception, which can affect the fidelity of the simulation as well as the performance. To address this limitation, future studies will incorporate a virtual reality headset (Fig.~\ref{fig:schematic}), providing 3D visualisation. The studies will also include expert robotic surgeons, which can validate the metrics being used.

\section*{REFERENCES}

\renewcommand{\section}[2]{}

\bibliographystyle{IEEEtran}
\bibliography{IeeeBstCtl,HSMR2018}

\end{document}